%% file: aaai22.tex
\def\year{2022}\relax
\documentclass[letterpaper]{article} 
\usepackage{aaai22}  
\usepackage{times}  
\usepackage{helvet}  
\usepackage{courier}  
\usepackage[hyphens]{url}  
\usepackage{graphicx} 
\usepackage{natbib}  
\usepackage{caption} 
\DeclareCaptionStyle{ruled}{labelfont=normalfont,labelsep=colon,strut=off} 
\usepackage[ruled,noline,algo2e]{algorithm2e}
\usepackage{algorithm}
\usepackage{url}            
\usepackage{booktabs}       
\usepackage{amsfonts}       
\usepackage{nicefrac}       

\usepackage{amsmath}
\SetKwBlock{DummyBlock}{}{}
\usepackage[hang,flushmargin]{footmisc}

\usepackage{newfloat}
\usepackage{listings}
\floatstyle{ruled}
\newfloat{listing}{tb}{lst}{}
\floatname{listing}{Listing}
\title{Towards Multi-Objective Statistically Fair Federated Learning}
\author{
   Ninareh Mehrabi\textsuperscript{\rm 1}\thanks{Work was done when Ninareh Mehrabi was an intern at Amazon Alexa AI. This work was accepted at AAAI 2022 workshop on Trustable, Verifiable and Auditable Federated Learning.}, Cyprien de Lichy\textsuperscript{\rm 2}, John McKay\textsuperscript{\rm 2}, Cynthia He\textsuperscript{\rm 2}, William Campbell\textsuperscript{\rm 2}
}
\affiliations{
    \textsuperscript{\rm 1}Information Sciences Institute, University of Southern California \\
    \textsuperscript{\rm 2}Amazon, Alexa AI \\
   \texttt{ninarehm@usc.edu}, \texttt{\{cllichy,jomcky,xih,cmpw\}@amazon.com}
}

\usepackage{bibentry}

\begin{document}

\maketitle

\begin{abstract}
\input{sections/1-abstract.tex}
\end{abstract}

\input{sections/2-introduction.tex}
\input{sections/3-method.tex}
\input{sections/4-experiments-results.tex}
\input{sections/5-related-work.tex}
\input{sections/6-conclusion.tex}

\bibliography{aaai22}
\clearpage

\end{document}

%% file: sections/1-abstract.tex
Federated Learning (FL) has emerged as a result of data ownership and privacy concerns to prevent data from being shared between multiple parties included in a training procedure. Although issues, such as privacy, have gained significant attention in this domain, not much attention has been given to satisfying statistical fairness measures in the FL setting. With this goal in mind, we conduct studies to show that FL is able to satisfy different fairness metrics under different data regimes consisting of different types of clients. More specifically, uncooperative or adversarial clients might contaminate the global FL model by injecting biased or poisoned models due to existing biases in their training datasets. Those biases might be a result of imbalanced training set~\cite{zhang2019fairness}, historical biases~\cite{10.1145/3457607}, or poisoned data-points from data poisoning attacks against fairness~\cite{Mehrabi_Naveed_Morstatter_Galstyan_2021,solans2020poisoning}. Thus, we propose a new FL framework that is able to satisfy multiple objectives including various statistical fairness metrics. Through experimentation, we then show the effectiveness of this method comparing it with various baselines, its ability in satisfying different objectives collectively and individually, and its ability in identifying uncooperative or adversarial clients and down-weighing their effect.

%% file: sections/2-introduction.tex
\section{Introduction}
Federated Learning (FL) has recently gained significant attention as a learning paradigm in which a central server orchestrates different clients and aggregates their models to obtain a global federated model. In this setup, the server has no access to the clients' data. Thus, the data remains local to the clients on which clients will train their own models over. In each round, the server will send a global model to the clients. Clients will receive the model and will train it further on their own local datasets. They will then send their models back to the server, and the server will aggregate these results in a secure manner. The server will use this new model for the next iteration as the initial model to be sent to each client.

Research has shown that such a learning framework can be beneficial to preserve privacy of the data \cite{thakkar2020understanding}. This along with other advantages that FL provides, including but not limited to not requiring the data to be transferred between many parties, has made FL a recent popular topic of interest. With the advancement of research and widespread interest in federated learning, different approaches have been proposed for private \cite{truong2020privacy}, personalized \cite{NEURIPS2020_24389bfe}, and fair \cite{DBLP:conf/iclr/LiSBS20,mohri2019agnostic} federated learning. Although methods have been proposed under fair FL literature by trying to make clients have uniform test accuracies~\cite{DBLP:conf/iclr/LiSBS20,mohri2019agnostic}, not much attention has been given to standard statistical fairness definitions. Amongst the ones that consider statistical fairness metrics \cite{cui2021addressing}, existence of uncooperative or adversarial clients have not been considered. Specifically, methods that rely on clients to locally satisfy fairness metrics~\cite{cui2021addressing} can be ineffective or unreliable in the existence of adversarial clients.

With these goals in mind, we propose a framework that is able to satisfy different objectives including but not limited to different fairness objectives. This framework is also able to identify uncooperative or adversarial clients who might inject poisoned, unfair, or poor quality models to the overall FL system~\cite{Mehrabi_Naveed_Morstatter_Galstyan_2021,solans2020poisoning}. This framework can also be considered as a verification or auditing framework in FL setups. In addition, we conduct studies to verify how this framework will work to satisfy different statistical fairness metrics under different conditions with the existence of different clients and compare this method against baselines, such as Federated Averaging (FedAvg) \cite{mcmahan2017communication}, Agnostic Federated Learning (AFL) \cite{mohri2019agnostic}, FCFL~\cite{cui2021addressing}, and other baselines from the fair FL literature \cite{DBLP:conf/iclr/LiSBS20}.

To summarize, in this work we aim to answer the following questions:
\begin{enumerate}
    \item In federated learning setup, where data is local to the clients and access to the sensitive attributes is considered to be a challenge, is it possible to satisfy different statistical fairness metrics, audit, and verify clients' models?
    \item Is it possible to satisfy multiple objectives including statistical fairness metrics using federated learning?
    \item Can we identify and mitigate the effect of uncooperative or adversarial clients who might inject malicious, unfair, and in general poor quality models into the federated learning system and instead reward better clients?
\end{enumerate}


%% file: sections/3-method.tex
\section{Methodology}

\subsection{Background}
FL has a client-server setup in which server's job is to orchestrate existing clients who each train local models on their own local datasets. These different trained models are then aggregated by the server in a new overall federated model. This aggregated model will then be used by the server for further training rounds which will be sent to the clients. The objective of the FL framework in this setup can be written as:

\begin{equation}
\underset{w}{\text{min}} f(w)= \sum_{k=1}^K p_k F_k(w) \quad \text{where} \quad F_k(w) = \frac{1}{n_k} \sum_{ik=1}^{n_k}f_{ik}(w)
\label{eqn:flobjective}
\end{equation}

Where $F_k$ is the local objective for client $k$, $f_{ik}(w)$ is the loss on example $i$ of client $k$ using model parameters $w$. Notice that one can solve this problem by considering $p_k$ as the probability that client $k$'s model will be incorporated into the federated model during the aggregation process by the server. For instance, in FedAvg \cite{mcmahan2017communication} $ p_k = \frac{n_k}{n}$, where $n_k$ is the number of examples in client $k$'s local dataset and $n$ is the total number of examples. 

\begin{algorithm}[t]
\SetAlgoLined
Input: $k$ number of clients; $\gamma_{j}$ weight for each objective $j$; $B$ local minibatch size; $E$ number of local epochs; $\eta$ learning rate.\\
Output: $w$ final federated model.\\
\textbf{Server Side:} 
\SetAlgoNoLine\DummyBlock{\SetAlgoLined
initialize $w_0$ \\
\For{t= 1,2,...}{
\For{each client $k$ in parallel}{
 $w_{t+1}^k$ $\leftarrow$ ClientUpdate($k$,$w_t$)  \\
 Validate client $k$'s model $w_{t+1}^k$ when temporarily aggregated with the global FL model $w_t$ and calculate $\sum_{j=1}^J \gamma_{j}s_{jk}.$ \\
}
Rank each client $k$ based on their scores $\sum_{j=1}^J \gamma_{j}s_{jk}$ and assign the rank score $rs_k$ to each client $k$. (//Optional step refer to the Ranking Algorithm for more details). \\ 
$w_{t+1}$ $\leftarrow$ $\sum_{k=1}^K \frac{rs_k}{\sum_{k=1}^Krs_k} w_{t+1}^k$ (//$rs_k$ can be replaced by $\sum_{j=1}^J \gamma_{j}s_{jk}$ if the optional step is skipped.) \\
}
\Return  $w_{t+1}$ \\}
\textbf{Client Side:} 
\SetAlgoNoLine\DummyBlock{\SetAlgoLined
ClientUpdate($k$,$w$): 
\SetAlgoNoLine\DummyBlock{\SetAlgoLined
\For{each local epoch i from 1 to $E$}{
\For{each batch $b$ with size $B$}{$w$ $\leftarrow$ $w - \eta \nabla \ell(w;b)$}
}
\Return $w$ to the server}}
\caption{FedVal Algorithm}
\label{fedval_algo}
\end{algorithm}

\begin{algorithm}[t]
\SetAlgoLined
\caption{Ranking Algorithm}
Input: Initial Step $\mu$; Step Size $\rho$. \\
Output: Clients' rank scores $rs_k$. \\
Initialize $rs_k$ with zeros for the first round; otherwise, use $rs_k$ scores from the previous rounds. \\
Sorted $\leftarrow$ Sort the clients based on their obtained scores $\sum_{j=1}^J \gamma_{j}s_{jk}$ from the FedVal algorithm. \\
\For{each client $k$ in Sorted}{
$rs_k$ += $\mu$ \\
$\mu$ $\times=$ $\rho$
}
\Return $rs_k$
\label{ranking}
\end{algorithm}
\subsection{FedVal}
To be able to satisfy different objectives including existing statistical fairness measures and be able to detect and down-weight the effect of uncooperative or adversarial clients who might train their models on imbalanced~\cite{zhang2019fairness}, poisoned~\cite{Mehrabi_Naveed_Morstatter_Galstyan_2021,solans2020poisoning}, or poor quality data~\cite{10.1145/3457607} that can contribute to the unfairness of their models, we decided to dedicate a central validation set for the server using which the server can then assign scores to the clients for validation purposes. This validation or verification step has a couple of advantages: (1) it gives the server a dataset on which it can compute fairness measures with regards to existing sensitive attributes in the data that can be used for auditing client models, (2) the server can compute scores for each client model and weight each client accordingly, (3) the validation set can audit the FL model with regards to any sensitive attribute a practitioner would have in mind making this framework flexible. Notice that in this setup, we assume that the server is a trustworthy party that plays a role in auditing the FL model and has the responsibility of assuring that the global model is a reliable model; however, we do not have the same trust toward clients and assume that there might exist adversarial or uncooperative clients in the system.

Although the idea of having a validation set in FL setup is not new \cite{stripelis2020accelerating}, our framework and aggregation mechanism are different. In addition, our work is focused on analyzing whether or not FL can satisfy and/or find a compromise between different fairness definitions which is an under-explored direction especially when considering the existence of uncooperative or adversarial clients.

To this end, we propose to use the same objective as in \ref{eqn:flobjective} except now $p_k = \frac {\sum_{j=1}^J \gamma_{j}s_{jk}} {\Gamma S}$ where $s_{jk}$ is the score that model from client $k$ has obtained for objective $j$ and $\gamma_j$ corresponds to the weight of objective $j$. $\Gamma S$ is the normalizing factor. This way, the model can find a compromise for multiple objectives, it can work with measures that need access to some data, such as fairness measures that require access to sensitive attributes. Moreover, it allows to identify uncooperative or adversarial clients through validation and verification. With this objective in mind, we write our federated learning algorithm as shown in Algorithm \ref{fedval_algo}.

As shown in Algorithm \ref{fedval_algo}, in each training round, the server sends a model to clients to train on their own dataset. The server will then fetch the clients' models and validate their contribution to the overall FL model and calculate the validation scores for each client. The clients' models will then be aggregated using a weighted average based on the validation scores. In this setting, we are trying to choose the best set of clients using our validation data. Thus, in a sense we are auditing each of the models trained by each client.

In order to make the client scoring more uniform and controllable, we also propose the optional ranking step that is explained in more details in Algorithm \ref{ranking}. This gives the server more control on assigning rewards to cooperative clients or punishing uncooperative clients by the budget that the validator server decides. Here, we propose one way of doing this ranking as an additional and optional step.

\subsection{Obtaining a validation set}

The choice of the validation set can be a challenge. Here we provide some options through which the central server can obtain a validation set in practice to perform the audits. The server could: 

1. Use historical data that the centralized model used for training before the existence of the FL framework as a validation data to validate the clients.  A downside to this can be that historical data might get outdated after a while, so the server might need to get a new type of data for its validation. However, with the existence of the federated framework, it would be harder for the server to obtain such data. 

2. Use publicly available data. A downside to this can be that public data might not well represent the clients in mind. 

3. Ask clients to donate some of their data for validation only. The downside of this would be that some of the clients' data would now be shared which is at odds with the federated learning framework where data stays local to each client and is not shared. However, such an approach would have lesser privacy concerns than typical centralized learning since only a small subset of clients' data could be shared for validation purposes, which might be acceptable in some settings. 

4. Make clients to become trainers and validators. Each client's data can be split into train and validation sets. A similar approach was proposed in \cite{stripelis2020accelerating}. A downside to this approach is that it can be costly as all clients need to both train and validate all the other clients. In addition, some clients may be adversarial and assign low or inaccurate scores to other clients.

5. Dedicate certain clients to only be validators and use their data for validation only to avoid the cost of the previous suggestion. This can bring down the cost issue as not all the clients need to both train and validate. In this case, we have dedicated trainers and validators. However, we still have the other remaining downside: the existence of adversarial clients among validators.

However, notice that despite the aforementioned challenges, having a validation set can have numerous advantages as previously mentioned. It can also give flexibility to the practitioner who audits such models as they can use and curate validation data that is appropriate for the use-case they have in mind considering demographic groups that might be more susceptible to be targets of a biased FL model depending on where the FL model is being deployed. 

%% file: sections/4-experiments-results.tex
\section{Experiments and Results}
We conduct two major studies to first demonstrate the effectiveness of the FedVal algorithm when compared to different existing FL algorithms and also its ability to satisfy multiple objectives. We then conduct studies to demonstrate the effect of having different ratios of cooperative vs uncooperative clients in the FedVal algorithm.  In the following sections, we are going to discuss the experiments we performed including the datasets used and the results.

\subsection{Datasets}
We conducted our experiments on two datasets: the UCI Adult dataset \footnote{https://archive.ics.uci.edu/ml/datasets/adult} which contains census data. The prediction task is whether or not an individual's income exceeds $50k$, gender (male or female) is the sensitive attribute, and the Heritage Health dataset \footnote{https://www.kaggle.com/c/hhp} which contains patient information and the task is to predict the Charleson Index (a survival indicator). We used age as a binarized sensitive attribute ($>45$ vs. $\leq 45$ years old).

\subsection{Metrics}
We report our results using accuracy as a performance metric as well as Statistical Parity Difference (SPD) \cite{10.1145/2090236.2090255} and Equality of Opportunity Difference (EOD) \cite{NIPS2016_9d268236} measures which are widely known statistical fairness metrics defined as follows:
\[ SPD = |p(\hat{Y}=+1|x \in \mathcal{D}_a)-p(\hat{Y}=+1|x \in \mathcal{D}_d)| \]
Where $D_a$ represents the advantaged demographic group and $D_d$ the disadvantaged group. SPD captures the differences between (advantaged and disadvantaged) demographic groups in getting assigned the positive outcome.

\begin{align*}
    EOD = |p(\hat{Y}=+1|x \in \mathcal{D}_a, Y=+1) \\ -p(\hat{Y}=+1|x \in \mathcal{D}_d, Y=+1)|
\end{align*}
EOD captures differences in the true positive rates among different (advantaged and disadvantaged) demographic groups.

\begin{figure*}[t]
\includegraphics[width=0.5\textwidth,trim=2cm 3cm 7cm 3cm,clip=true]{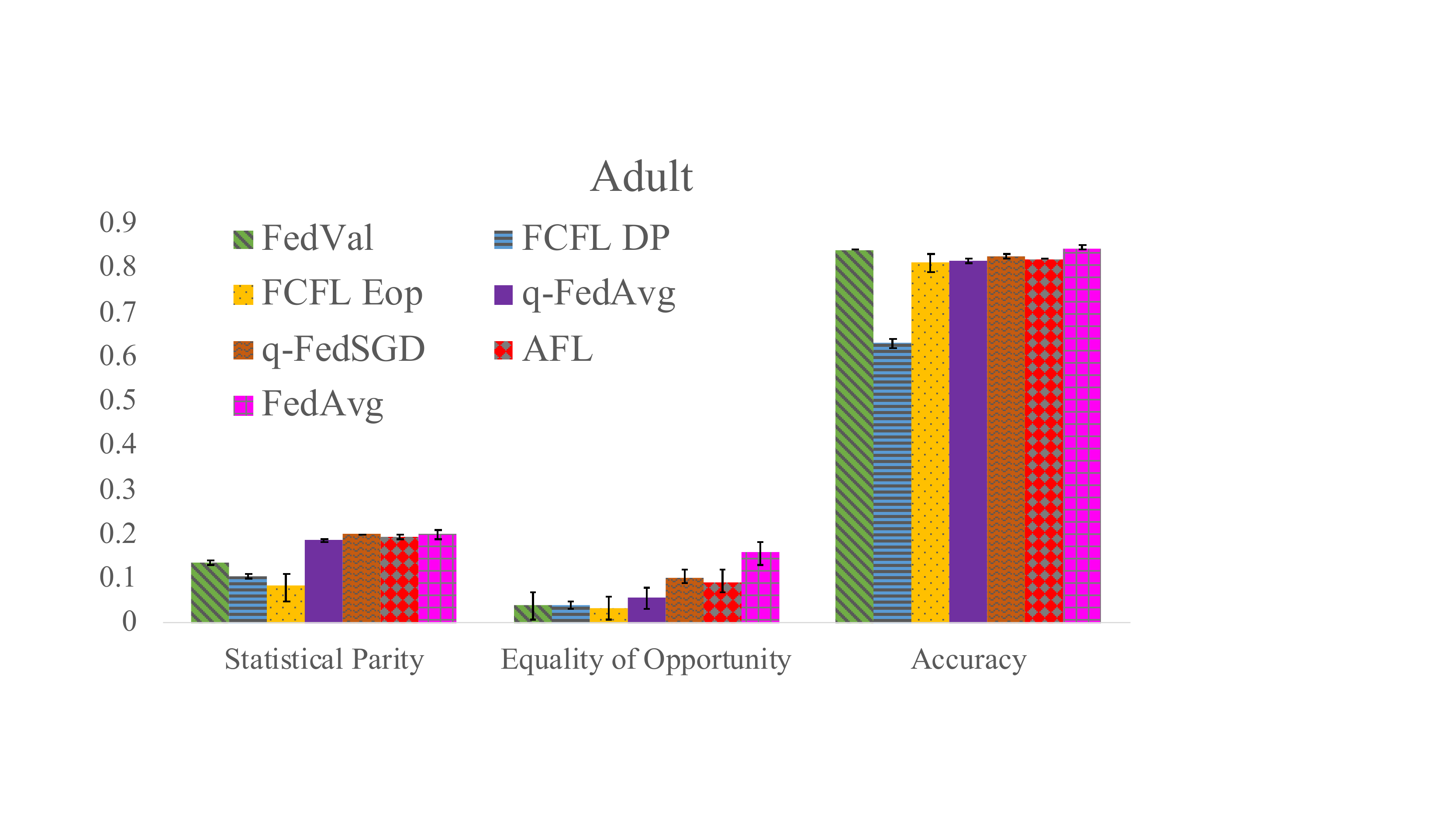}
\includegraphics[width=0.5\textwidth,trim=2cm 3cm 7cm 3cm,clip=true]{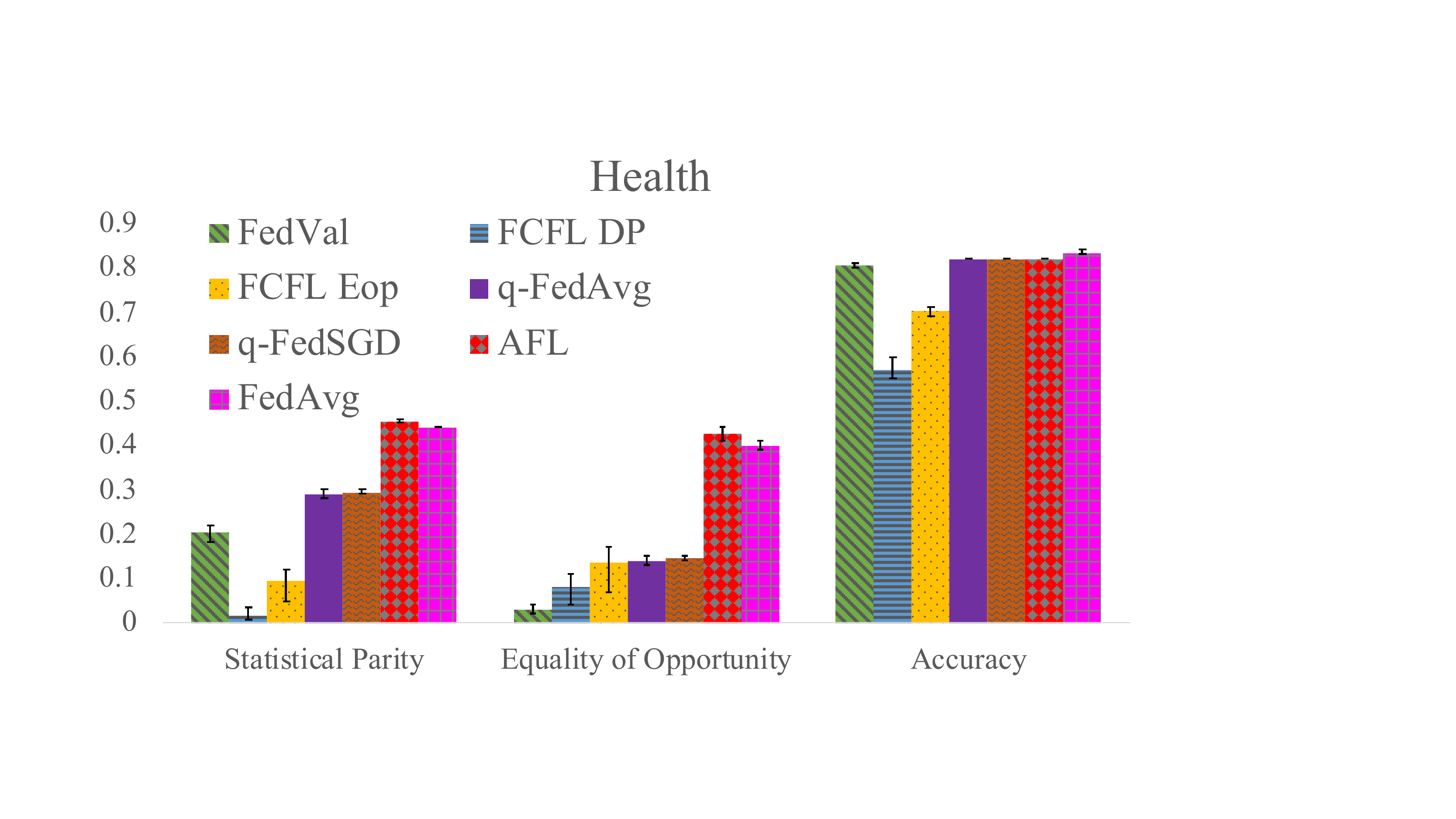}
\caption{FedVal results compared to different baseline FL algorithms. FedVal is shown to be able to maintain a good balance in satisfying all the three objectives collectively without sacrificing accuracy for the price of fairness.}
\label{fig1:baselines}
\end{figure*}

\begin{figure*}[t]
\includegraphics[width=0.24\textwidth,trim=6cm 5cm 14cm 5cm,clip=true]{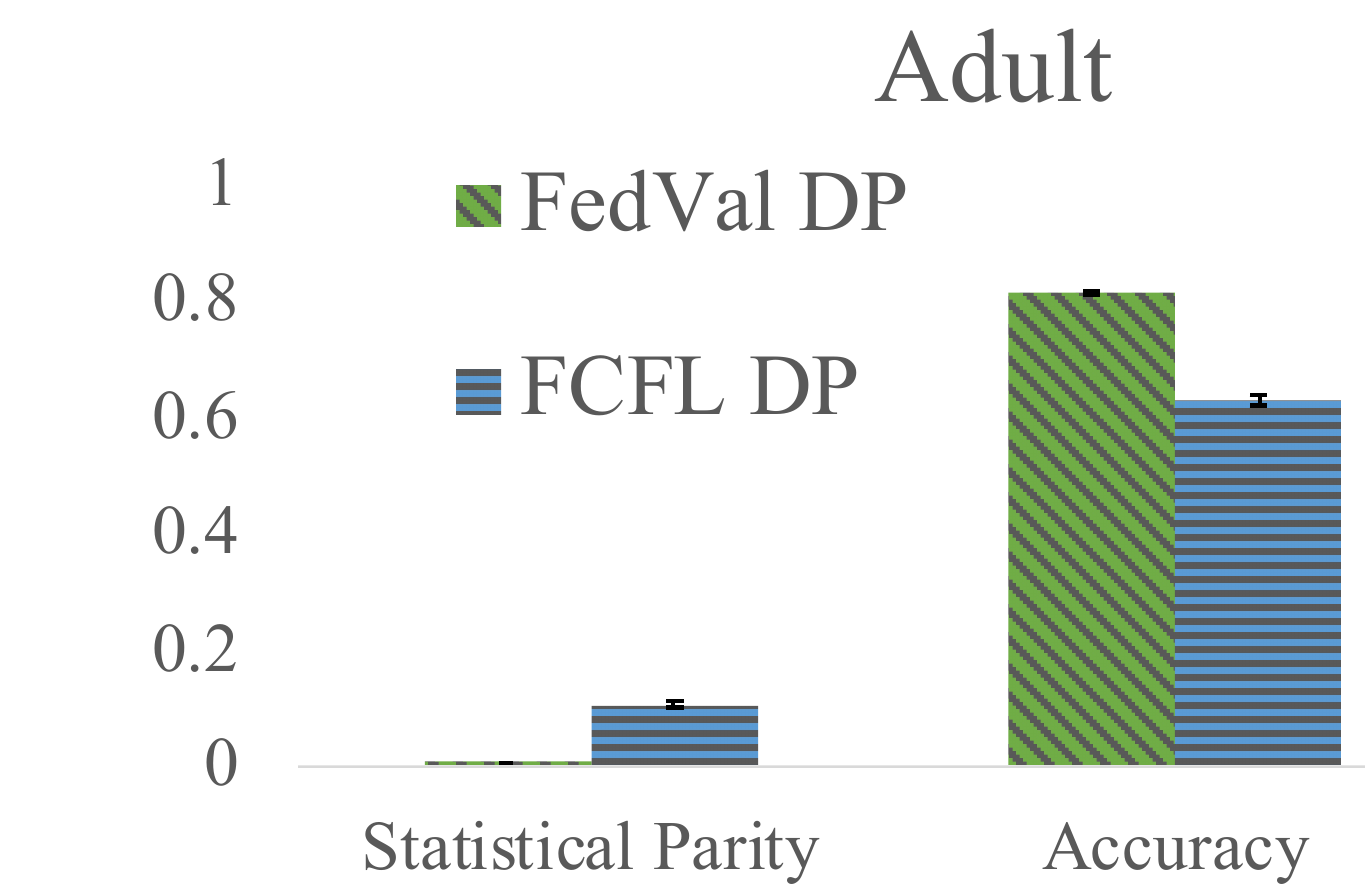}
\includegraphics[width=0.24\textwidth,trim=6cm 5cm 14cm 5cm,clip=true]{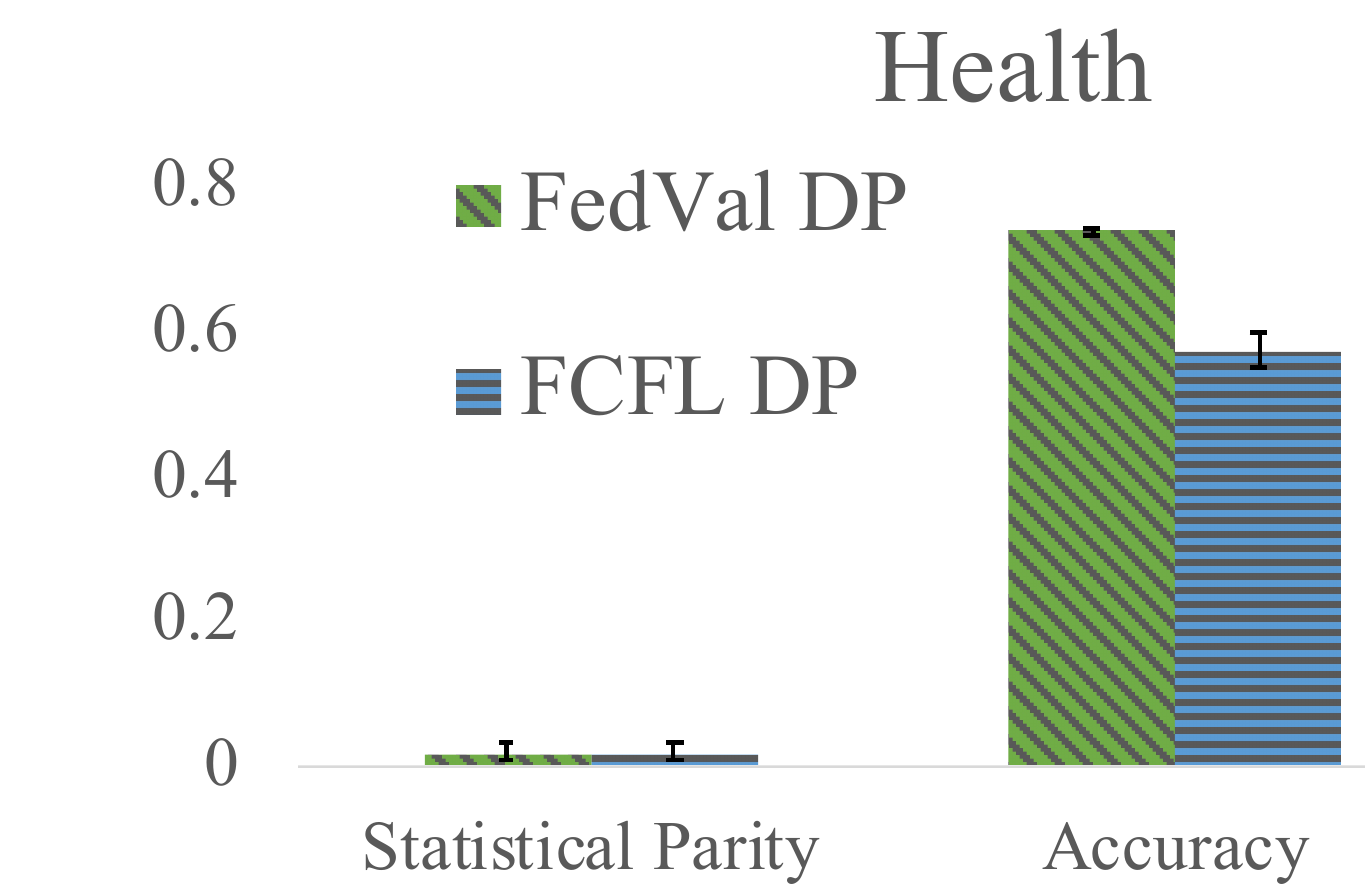}
\includegraphics[width=0.24\textwidth,trim=6cm 5cm 14cm 5cm,clip=true]{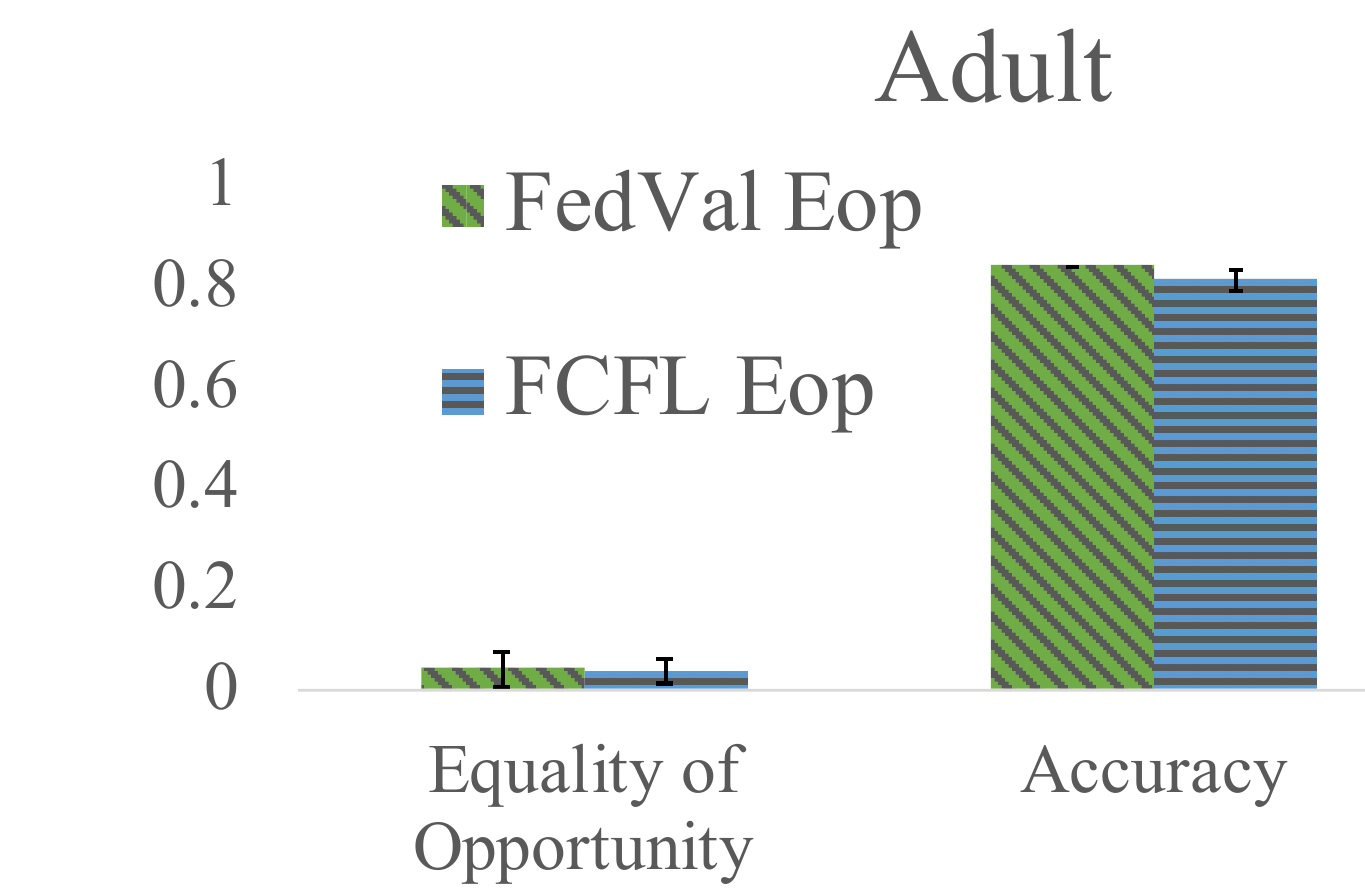}
\includegraphics[width=0.24\textwidth,trim=6cm 5cm 14cm 5cm,clip=true]{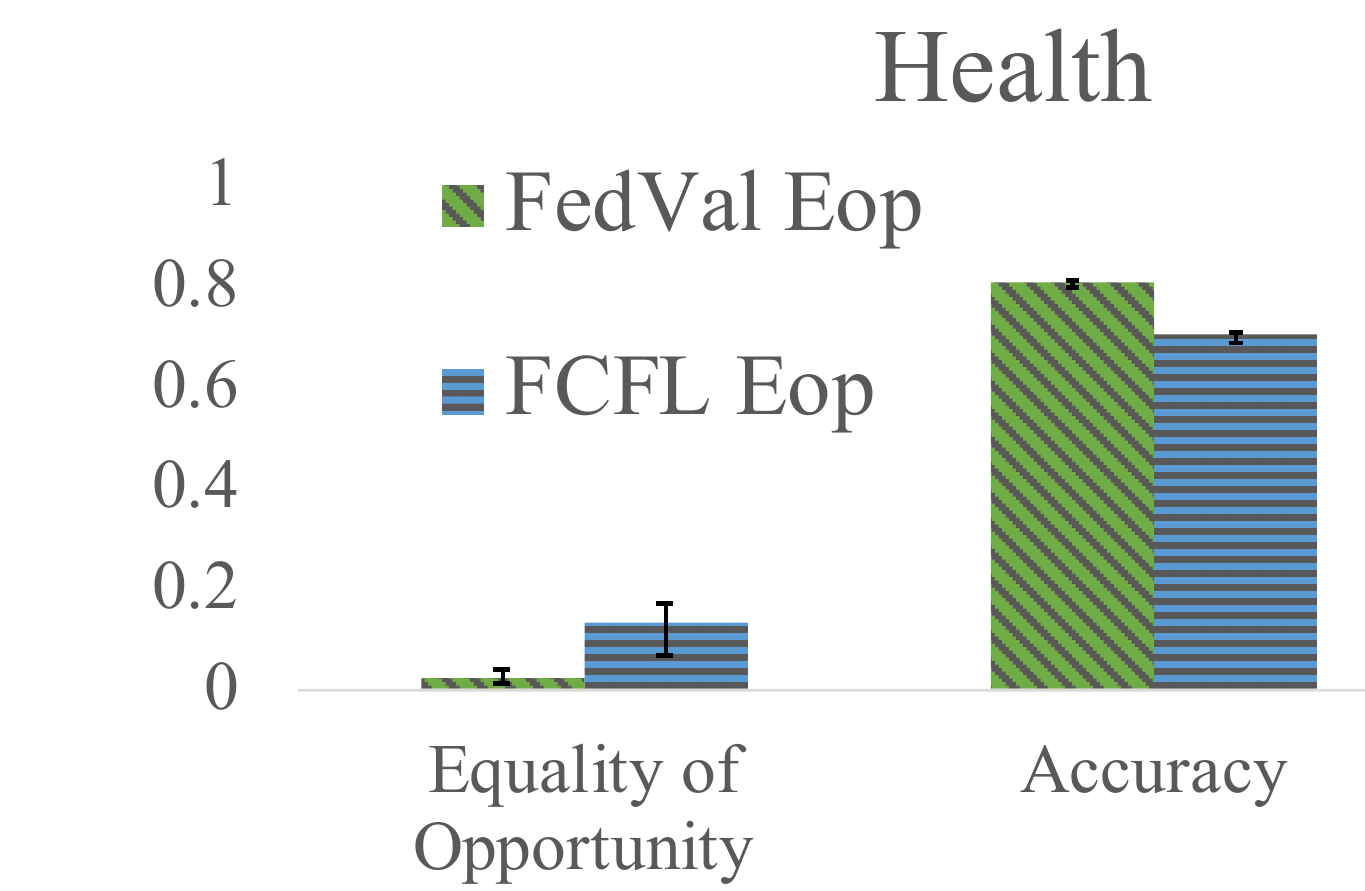}
\caption{Results comparing FCFL optimized for a specific fairness objective vs. FedVal optimized for the same corresponding fairness objective. We observe that FedVal is able to obtain more fair results by having lower SPD and EOD while maintaining higher accuracy values compared to FCFL. Lower Statistical Parity and Equality of Opportunity differences represent lower bias with regards to these two fairness measures; thus, higher fairness and better results.}
\label{fig:fclvsfedval}
\end{figure*}

\begin{figure*}[t]
\includegraphics[width=0.5\textwidth,trim=1cm 3cm 10cm 3cm,clip=true]{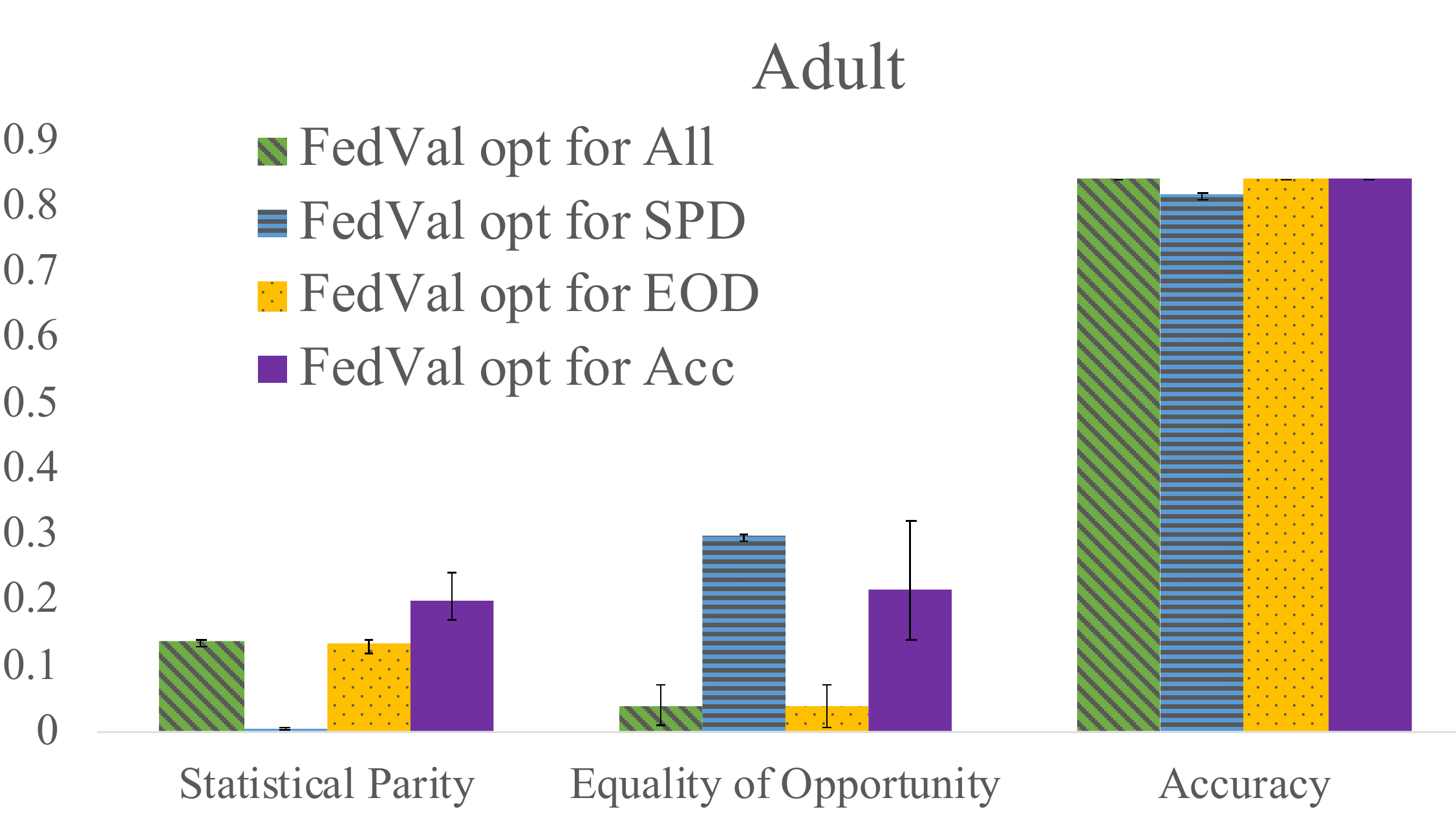}
\includegraphics[width=0.5\textwidth,trim=1cm 3cm 10cm 3cm,clip=true]{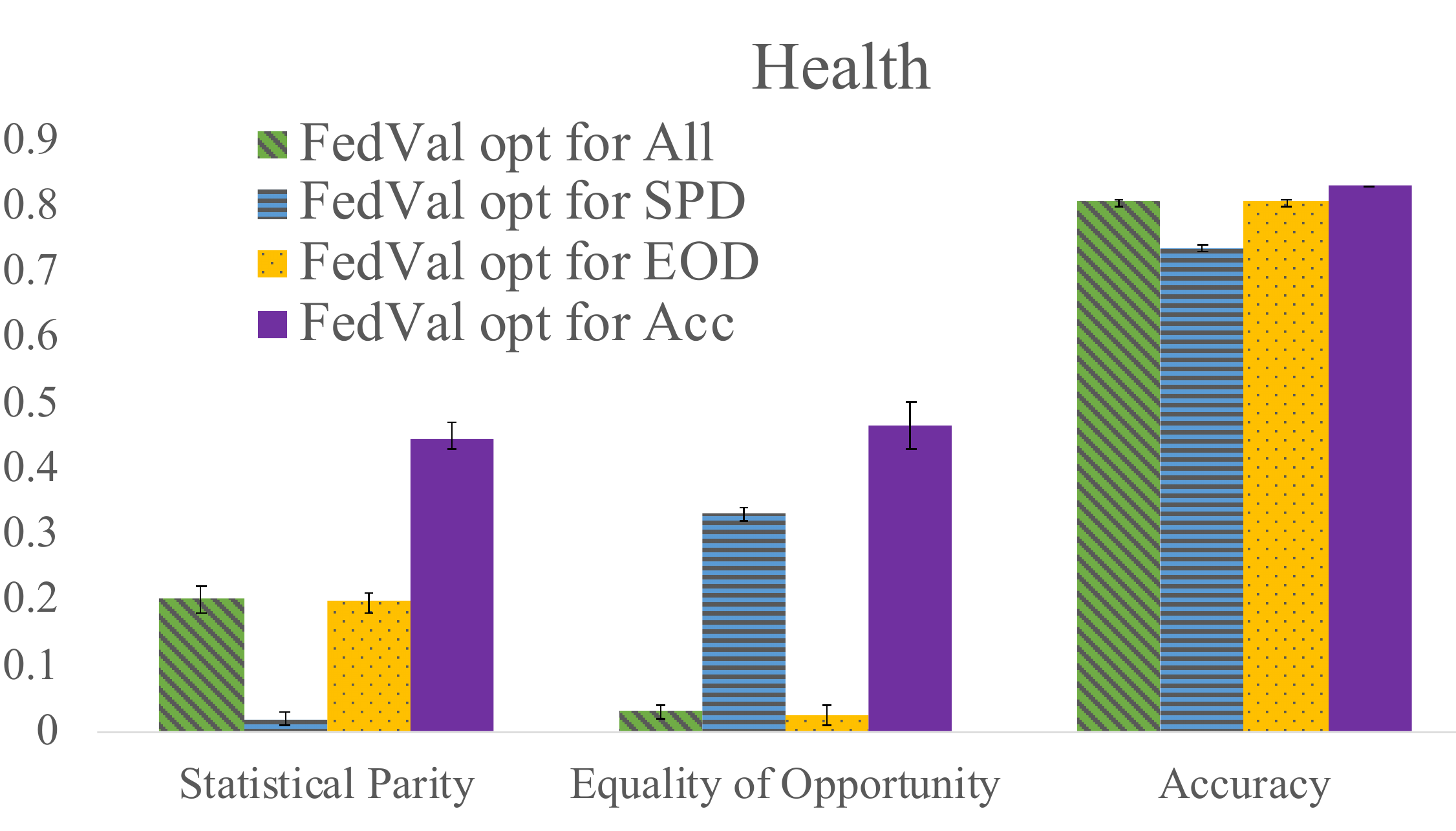}
\caption{FedVal results for when it satisfies all the three objectives collectively vs. each objective individually. Lower Statistical Parity and Equality of Opportunity differences represent lower bias with regards to these two fairness measures; thus, higher fairness and better results.}
\label{fig2:objectives}
\end{figure*}
\subsection{Verifying FedVal}
To verify the FedVal algorithm, we considered two aspects: (1) Verifying the effectiveness of FedVal compared to different existing FL algorithms, (2) Testing FedVal's ability in satisfying different objectives including different statistical fairness metrics along with accuracy. To put FedVal to the test, we needed different varieties of clients, the cooperative clients who would train less biased models, the uncooperative clients who would intentionally train their models on skewed/imbalanced data to bias their trained models~\cite{Mehrabi_Naveed_Morstatter_Galstyan_2021,solans2020poisoning,zhang2019fairness} and normal clients. To satisfy these objectives, for this set of experiments, we created 10 clients with different varieties and evaluated FedVal accordingly.
\subsubsection{Against Baselines}
We compared FedVal against baselines in its ability to satisfy three objectives: fairness objectives such as Statistical Parity Difference (SPD) \cite{10.1145/2090236.2090255}, Equality of Opportunity Difference (EOD) \cite{NIPS2016_9d268236} and accuracy. The baselines considered were FedAvg \cite{mcmahan2017communication}, AFL \cite{mohri2019agnostic}, q-FedAvg \cite{DBLP:conf/iclr/LiSBS20}, q-FedSGD \cite{DBLP:conf/iclr/LiSBS20}, and FCFL~\cite{cui2021addressing}. In this section, we set up FedVal to optimize for the three objectives, namely SPD, EOD, and accuracy, collectively. In a follow-up experiment, we show results on FedVal satisfying each of these objectives individually. For the baselines, we used standard FL algorithms as well as algorithms designed for fairness in the FL setup. \textit{\textbf{FedAvg}} is a widely known standard FL algorithm. \textit{\textbf{AFL}} is a FL algorithm designed to satisfy fairness in FL setup based on the notion of \textit{good-intent fairness} in which the goal is to minimize the maximum loss obtained by any protected class/client. In simpler words, the goal in \textit{\textbf{AFL}} is to maximize the performance of the worst agent. \textit{\textbf{q-FedAvg}} and \textit{\textbf{q-FedSGD}} are algorithms also specifically designed to obtain fairness in FL setting in which the goal is for clients to obtain more uniform accuracy. Finally, \textit{\textbf{FCFL DP}} and \textit{\textbf{FCFL Eop}} which are recent FL algorithms     designed to satisfy statistical fairness objectives in which \textit{\textbf{FCFL DP}} is optimized for SPD specifically and \textit{\textbf{FCFL Eop}} fro EOD. We report the averaged results over three different data splits along with error bars in Fig \ref{fig1:baselines}. 

\subsubsection{Results}
The results in Fig \ref{fig1:baselines} demonstrate that FedVal is able to effectively compromise for different objectives by obtaining a balanced results in satisfying fairness measures as well as accuracy without sacrificing one objective over another. Although FCFL is able to obtain fair outcomes with regards to the objective that it is optimized for, notice that FCFL is sacrificing accuracy which is not the case for FedVal that is trying to obtain a balance between all the objectives collectively. We also show in Fig \ref{fig:fclvsfedval} that FedVal is able to outperform FCFL in satisfying fairness metrics when FedVal is set to satisfy the corresponding fairness measure that FCFL is specifically optimized for which is a more fair comparison to have. These results also demonstrate the ability of FedVal in effectively identifying poor quality models injected by the uncooperative or adversarial clients, diminishing their effects through validation, and obtaining more fair outcomes.

\subsubsection{Against Different Objectives}
In addition to baselines, we verified FedVal's ability in satisfying different objectives. We compared results when FedVal optimizes different objectives individually vs when it optimizes all the objectives collectively. This experiment also demonstrates the impact of varying the weights of the objectives in FedVal since not considering an objective in FedVal will be the equivalent to zeroing out the weight for the corresponding objective. We considered accuracy to test the general performance of the model along with Statistical Parity Difference (SPD) and Equality of Opportunity Difference (EOD) as standard fairness metrics as discussed and used in previous sections. Similar to the results reported in the previous section, we report the averaged results over three different data splits along with error bars in Fig \ref{fig2:objectives}. 

Results in Fig \ref{fig2:objectives} demonstrate that FedVal is able to satisfy each of the objectives including statistical fairness metrics individually. These results are nice additions to our previous results in which FedVal was shown to maintain a balance in satisfying different objectives collectively. 

\subsection{Fedval with Different Client Ratios}
In addition to verifying FedVal, we perform experiments here to demonstrate the effect of having different types of clients, namely uncooperative vs cooperative in our framework. With this goal in mind, we applied FedVal on different sets of datasets each having different ratios of cooperative clients. In these sets of experiments, we concentrate on fairness metrics specifically. Thus, we skew the datasets for certain clients to make them produce biased models trained on imbalanced datasets and call them the uncooperative clients who inject biased models vs. clients who would train their models on a balanced datasets and would inject more fair models compared to the uncooperative clients. Similar to the experiments performed in previous sections, we utilize SPD and EOD as our fairness measures. In addition, we report results having the optional ranking schema in our introduced algorithm compared to not having the specific ranking schema.

\begin{figure*}[t]
\includegraphics[width=0.5\textwidth,trim=11cm 6cm 11cm 6cm,clip=true]{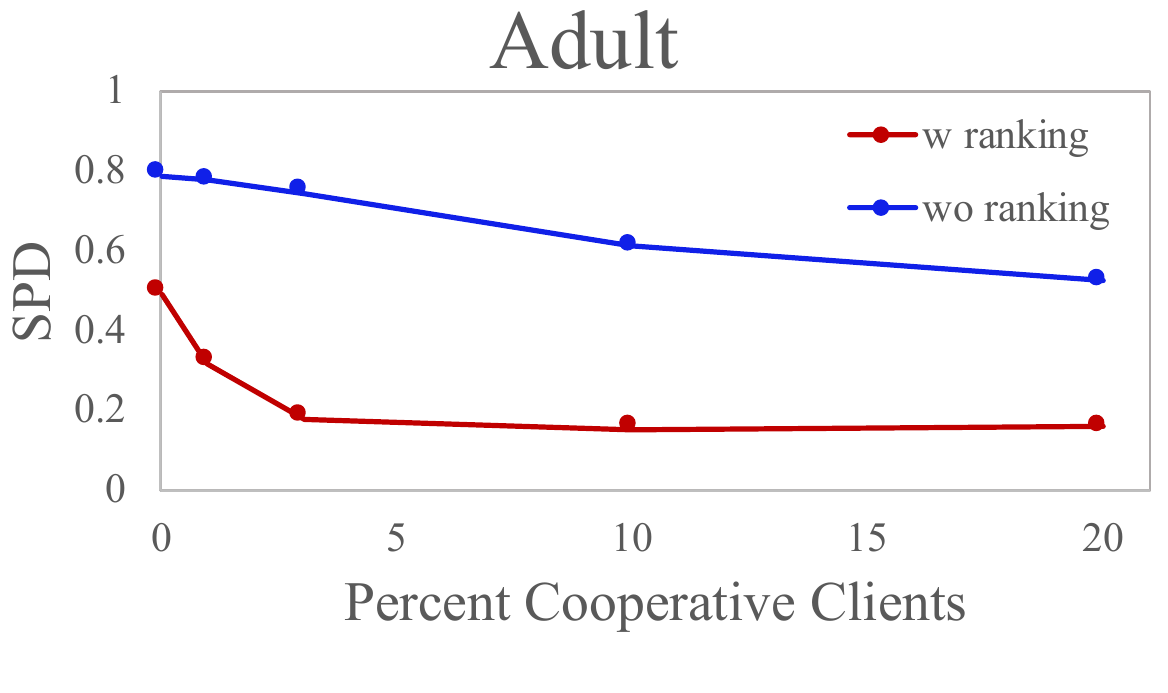}
\includegraphics[width=0.5\textwidth,trim=11cm 6cm 11cm 6cm,clip=true]{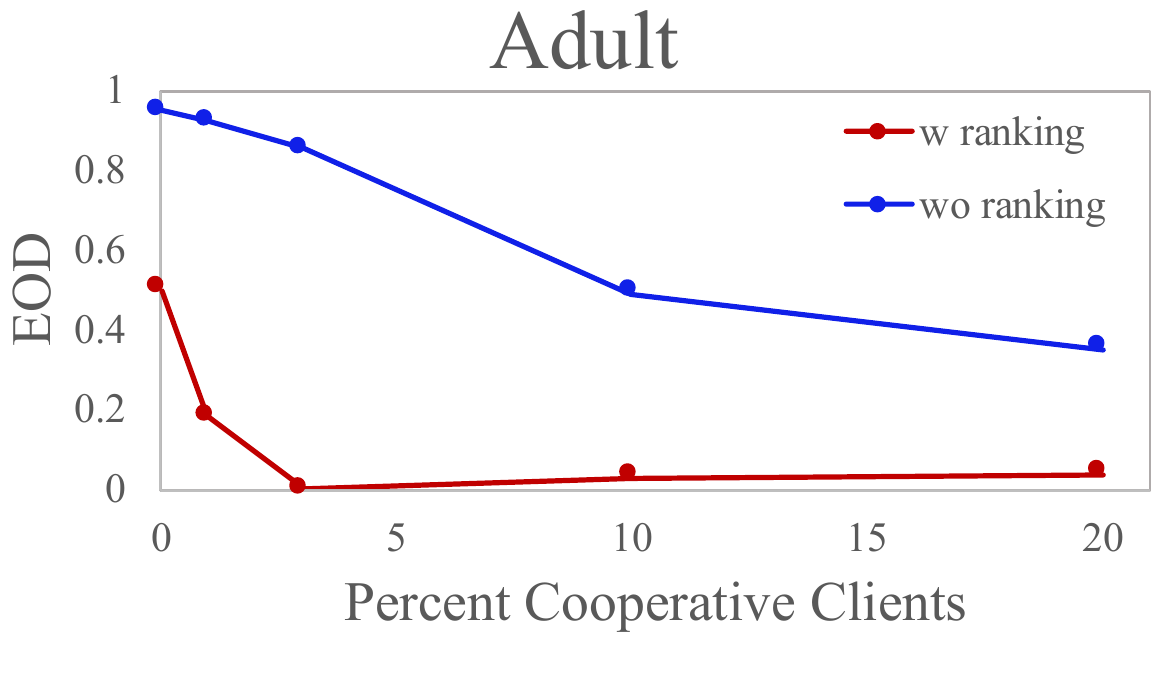}
\includegraphics[width=0.5\textwidth,trim=11cm 6cm 11cm 6cm,clip=true]{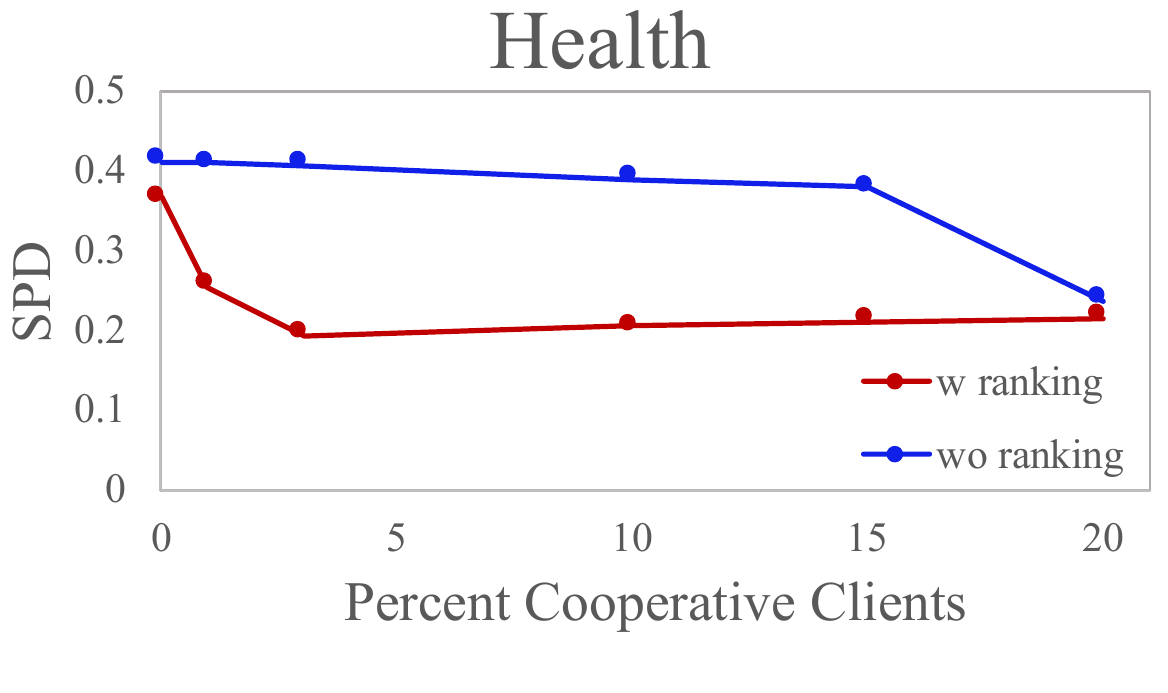}
\includegraphics[width=0.5\textwidth,trim=11cm 6cm 11cm 6cm,clip=true]{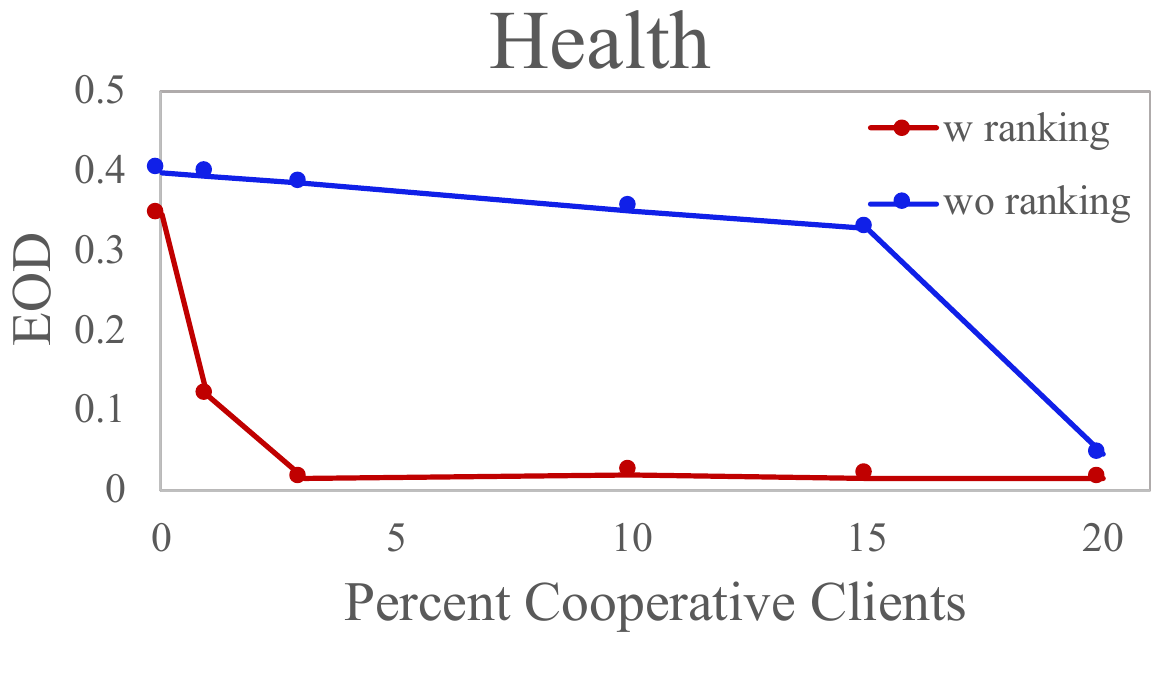}
\caption{Different cooperative client ratio results on Adult and Health according to SPD and EOD fairness measures.}
\label{fig3:ratios}
\end{figure*}

Results in Fig \ref{fig3:ratios} demonstrate that by increasing the number of cooperative clients who would inject more fair models compared to the unfair clients, the bias in terms of SPD and EOD will decrease as expected. In addition, this decrease is more noticeable when we use our introduced optional ranking schema. One can observe that in the ranking setup the significant decrease in bias happens around when we have ~3\% of cooperative users in both of the datasets compared to 10\%-15\% in the no ranking setup. This is because the ranking schema punishes the uncooperative users more harshly and rewards the cooperative users more aggressively compared to a setup where there is no ranking schema. Of course, one could design other ranking strategies and control this trade-off according to their use-case. These results once again verify the effectiveness of FedVal in identifying adversarial or uncooperative clients and its effectiveness in reducing their effects in the FL model. This demonstrates FedVal's verification, validation, and auditing capabilities.

%% file: sections/5-related-work.tex
\section{Related Work}

\subsection{Federated Learning}
Research in federated learning has expanded drastically in recent years \cite{kairouz2019advances}. Not only work has been done to optimize federated learning in general \cite{9155494,li2018federated}, but the research has expanded to other areas of research as well, such as privacy and federated learning \cite{truong2020privacy}, personalization and federated learning \cite{NEURIPS2020_24389bfe}, and more recently work has been done in the context of fairness and federated learning \cite{DBLP:conf/iclr/LiSBS20,mohri2019agnostic}. In addition, federated learning has gained significant importance in health care domain \cite{rieke2020future} and many other applications of machine learning. Not only in machine learning, but work has expanded to Natural Language Processing (NLP) as well \cite{lin2021fednlp}. This significance and penetration of federated learning into different applications, including sensitive applications that can affect our society, brings the need to think about fairness implications of federated learning. 

\subsection{Fairness}
Similar to federated learning, research in fair Machine Learning (ML) and Natural Language Processing (NLP) has gained significant popularity in recent years \cite{10.1145/3457607}. Different statistical fairness metrics have been proposed as measures for fairness, such as statistical parity \cite{10.1145/2090236.2090255} and equality of opportunity \cite{NIPS2016_9d268236}. We utilized some of these measures in our studies as well. However, work in fairness does not conclude itself in proposing measures. Researchers try to constantly make different existing algorithms and models more fair in different tasks and applications, such as classification \cite{pmlr-v54-zafar17a} and regression \cite{pmlr-v97-agarwal19d} in general ML, and many other NLP applications, such as translation \cite{basta-etal-2020-towards}, language generation \cite{liu-etal-2020-mitigating}, named entity recognition~\cite{10.1145/3372923.3404804}, and commonsense reasoning tasks~\cite{mehrabi2021lawyers}. Thus, we see federated learning algorithms no exception from being included in such studies considering its applications in various different sensitive environments, such as healthcare systems. 

\subsection{Fair Federated Learning}
Some previous work tackled the fairness problem in FL \cite{DBLP:conf/iclr/LiSBS20,mohri2019agnostic,hao2021towards,9378043,10.1145/3375627.3375840}. However, they mostly focused on making the FL setup fair for the participating clients and did not consider statistical fairness metrics. Amongst the ones that considered statistical fairness metrics~\cite{cui2021addressing}, they either considered satisfying such metrics locally by trusting the clients which might not be effective in case of having adversarial clients, or in general they did not consider cases where auditing and verification is needed, such as cases where the client data itself might be intentionally biased or poisoned~\cite{Mehrabi_Naveed_Morstatter_Galstyan_2021,solans2020poisoning} and can corrupt the final global FL model. Thus, although most of the previous work in fair federated learning focused on having a framework in which clients with different data distributions can be treated fairly and similarly to each other, not much attention has been given to standard statistical fairness metrics with regards to the existing sensitive attributes in the data and the destructive outcomes the unfair FL model can have in the existence of adversarial, uncooperative, or unfair clients who can train unfair models by poisoning their data instances \cite{Mehrabi_Naveed_Morstatter_Galstyan_2021}. Even if the clients may not be adversarial, chances are that some clients may be training their models on an unintentionally biased data~\cite{10.1145/3457607} that can corrupt the overall FL model. That is one of the motivations for our work.

%% file: sections/6-conclusion.tex
\section{Conclusion}
In this work, we proposed a new simple yet effective FL framework, FedVal, that is able to satisfy multiple objectives including various fairness measures and compared it with other FL baseline algorithms. We analyzed FedVal's capability in satisfying statistical fairness metrics in different scenarios with varying ratios of uncooperative or adversarial clients. In addition, We showed that FedVal is able to reduce the bias introduced by uncooperative or adversarial clients. By including a validation step to rate clients, FedVal is able to achieve higher fairness. As a future direction, it would be interesting to add privacy and robustness objectives and analyze whether FedVal can satisfy those as well. In addition, one could investigate the incompatibility between fairness definitions \cite{kleinberg2016inherent} in the FL setting. For instance, in which settings one could reduce one type of bias at the expense of another fairness measure, in which settings is it possible to optimize all fairness objectives? It would also be interesting to compare and contrast these issues in FL vs centralized settings.